\documentclass[11pt]{article}
\usepackage{acl2010}
\usepackage{times}
\usepackage{latexsym}
\usepackage{booktabs} 
\usepackage{graphicx}
\usepackage{caption}
\usepackage{subcaption}
\usepackage{lineno}
\usepackage{multirow}
\usepackage{url}
\usepackage{amssymb}
\usepackage{amsmath}
\usepackage{fancyvrb}
\usepackage{listings}
\usepackage{float}
\usepackage{url}
\usepackage{tikz}
\usepackage{environ}
\usepackage[normalem]{ulem}
\useunder{\uline}{\ul}{}

\makeatletter
\newsavebox{\measure@tikzpicture}
\NewEnviron{scaletikzpicturetowidth}[1]{%
  \def\tikz@width{#1}%
  \begin{lrbox}{\measure@tikzpicture}%
  \BODY
  \end{lrbox}%
  \pgfmathparse{#1/\wd\measure@tikzpicture}%
  \BODY
}
\makeatother

\title{Utilizing Wordnets for Cognate Detection among Indian Languages}

\author{
\textbf{Diptesh  Kanojia}\textsuperscript{$\dagger$,$\clubsuit$,$\star$},  \textbf{Kevin Patel}\textsuperscript{$\dagger$}, \textbf{Pushpak Bhattacharyya}\textsuperscript{$\dagger$},\\ \textbf{Malhar Kulkarni}\textsuperscript{$\dagger$}, \textbf{Gholamreza Haffari}\textsuperscript{$\star$}\\
\textsuperscript{$\dagger$}Indian Institute of Technology Bombay, India\\
\textsuperscript{$\clubsuit$}IITB-Monash Research Academy, India\\
\textsuperscript{$\star$}Monash University, Australia\\
\textsuperscript{$\dagger$}\{diptesh, kevin, pb, malhar\}@iitb.ac.in\\
\textsuperscript{$\star$}gholamreza.haffari@monash.edu\\
}

\date{}

\begin{document}
\maketitle
\begin{abstract}
Automatic Cognate Detection (ACD) is a challenging task which has been utilized to help NLP applications like Machine Translation, Information Retrieval and Computational Phylogenetics. Unidentified cognate pairs can pose a challenge to these applications and result in a degradation of performance. In this paper, we detect cognate word pairs among ten Indian languages with Hindi and use deep learning methodologies to predict whether a word pair is cognate or not. We identify IndoWordnet as a potential resource to detect cognate word pairs based on orthographic similarity-based methods and train neural network models using the data obtained from it. We identify parallel corpora as another potential resource and perform the same experiments for them. 

We also validate the contribution of Wordnets through further experimentation and report improved performance of up to 26\%. We discuss the nuances of cognate detection among closely related Indian languages and release the lists of detected cognates as a dataset. We also observe the behaviour of, to an extent, unrelated Indian language pairs and release the lists of detected cognates among them as well.
\end{abstract}

\section{Introduction}
Cognates are words that have a common etymological origin \cite{crystal2008dictionary}. They account for a considerable amount of unique words in many lexical domains, notably technical texts. The orthographic similarity of cognates can be exploited in different tasks involving recognition of translational equivalence between words, such as machine translation and bilingual terminology compilation.  For \textit{e.g.,} the German - English cognates, \textit{Blume - bloom} can be identified as cognates with orthographic similarity methods. Detection of cognates helps various NLP applications like IR \cite{pranav2018alignment}. \newcite{rama2018automatic} study various cognate detection techniques and provide substantial proof that automatic cognate detection can help infer phylogenetic trees. In many NLP tasks, the orthographic similarity of cognates can compensate for the insufficiency of other kinds of evidence about the translational equivalency of words \cite{mulloni2006automatic}. The detection of cognates in compiling bilingual dictionaries has proven to be helpful in Machine Translation (MT), and Information Retrieval (IR) tasks \cite{meng2001generating}. Orthographic similarity-based methods have relied on the lexical similarity of word pairs and have been used extensively to detect cognates \cite{ciobanu2014automatic,mulloni2007automatic,inkpen2005automatic}. These methods, generally, calculate the similarity score between two words and use the result to build training data for further classification. Cognate detection can also be performed using phonetic features and researchers have previously used consonant class matching (CCM) \cite{turchin2010analyzing}, sound class-based alignment (SCA) \cite{list2010sca} \textit{etc.} to detect cognates in multilingual wordlists. The identification of cognates, here, is based on the comparison of words sound correspondences. Semantic similarity methods have also been deployed to detect cognates among word pairs \cite{kondrak2001identifying}. The measure of semantic similarity uses the context around both word pairs and helps in the identification of a cognate word pair by looking of similarity among the collected contexts. 

\begin{figure}[!ht]
\includegraphics[width=1\linewidth]{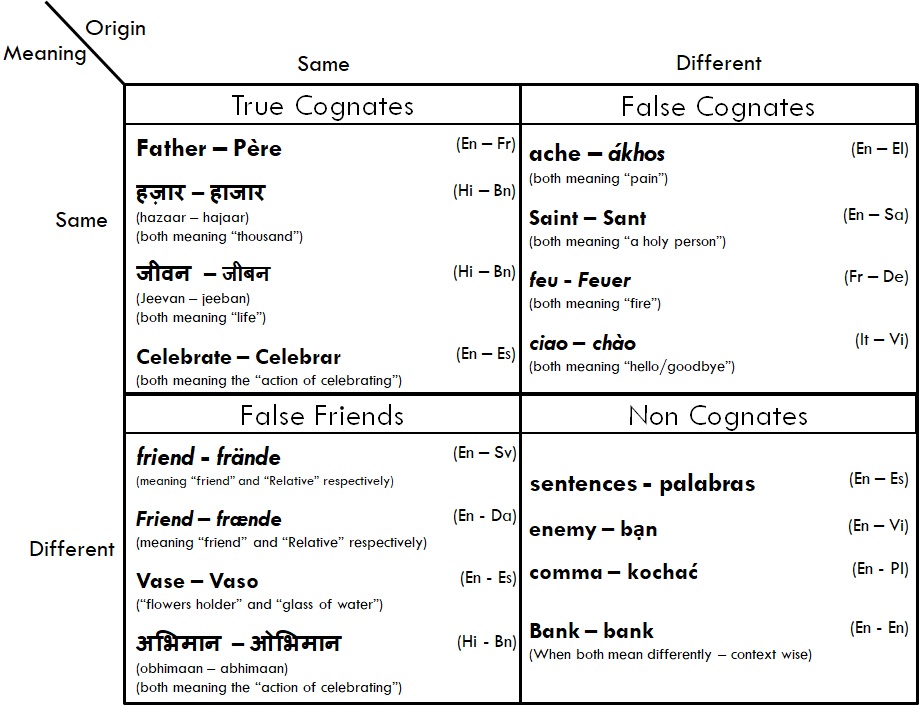}
\caption{The Cognate Identification Matrix}
\label{fig:cogmat}
\end{figure}

\begin{figure}[!ht]
\includegraphics[width=1\linewidth]{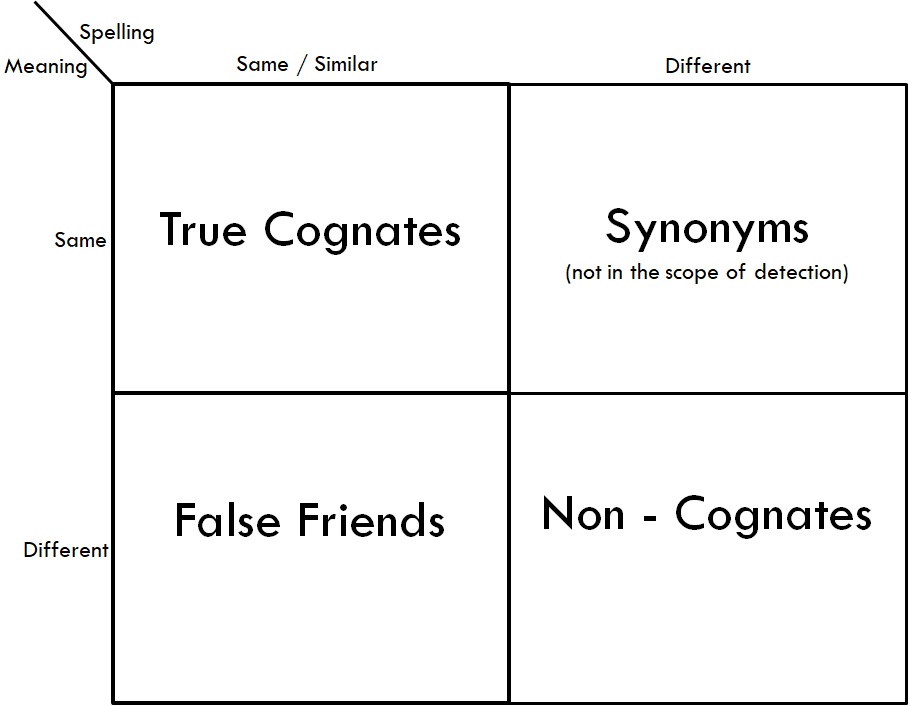}
\caption{Scope of our work; Detection of True Cognates and False Friends}
\label{fig:cogmat2}
\end{figure}

For our work, we can primarily divide words into four main categories \textit{viz.} \textbf{True Cognates, False Cognates, False Friends and Non-Cognates}. In Figure \ref{fig:cogmat}, we present this classification with examples from various languages along with their meanings for better understanding. While some false friends are also false cognates, most of them are genuine cognates. \textit{Our primary goal is to be able to identify True Cognates.} Sanskrit (Sa) is known to be the mother of most of the Indian languages. Hindi (Hi), Bengali (Bn), Punjabi (Pa), Marathi (Mr), Gujarati (Gu), Malayalam (Ml), Tamil (Ta) and Telugu (Te) are known to borrow many words from it. Thus, one may observe that \textit{words which belong to the same concept in these languages, if orthographically similar, are True Cognates.} Currently, we include loan words in the dataset used for our work and include them as cognates. Since, eventually we aim to apply our work to Machine Translation and other NLP applications, we believe that this would help establish a better correlation among source-target language pairs. Also, we do not detect false friends and hence restrict the scope of True cognate detection using this hypothesis to Figure \ref{fig:cogmat2}.

\textbf{We utilize the synset information from linked Wordnets to identify words within the same concept and deploy orthographic similarity-based methods to compute similarity scores between them.} This helps us identify words with a high similarity score. In case of most of the Indian languages, a sizeable contribution of words/concepts is loaned from the Sanskrit language. In linked IndoWordnet, each concept is aligned to the other based on an `id' which can be reliably used as a measure to say that the etymological origin is the same, for both the concepts. Hence, words with the same orthographic similarity can be said to be \textbf{`True Cognates'}. Using this methodology, we detect highly similar words and use them as training data to build models which can predict whether a word pair is cognate or not. The rest of the paper is organized as follows. In Section 2 we describe the related work that has been carried out on cognate detection together with some of its practical applications, while in Section 3 we present our approach and deal in greater detail with our learning algorithms. Once the proposed methodology has been outlined, we step through an evaluation method we devised and report on the results obtained as specified in Section 4. Section 5 concludes our paper with a brief summary and tackling further challenges in the near future.

\subsection{Contributions}

We make the following contributions in this paper:\\
1. We perform cognate detection for eleven Indian Languages.\\
2. We exploit Indian languages behaviour to obtain a list of true cognates (WNdata from WordNet and PCData from Parallel Corpora).\\
3. We train neural networks to establish a baseline for cognate detection.\\
4. We validate the importance of Wordnets as a resource to perform cognate detection.\\
5. We release our dataset (WNdata + PCdata) of cognate pairs publicly for the language pairs Hi - Mr, Hi - Pa, Hi - Gu, Hi - Bn,  Hi - Sa, Hi - Ml, Hi - Ta, Hi - Te, Hi - Ne, and Hi - Ur.

\section{Related Work}
\label{sec:RelWork}

One of the most common techniques to find cognates is based on the manual design of rules describing how orthography of a borrowed word should change, once it has been introduced into the other language. \newcite{koehn2000estimating} expand a list of English-German cognate words by applying well-established transformation rules. They also noted that the accuracy of their algorithm increased proportionally with the length of the word since the accidental coexistence of two words with the same spelling with different meanings (we identify them as `false friends') decreases the accuracy. 

Most previous studies on automatic cognate identification do not investigate Indian languages. Most of the Indian languages borrow cognates or ``loan words'' from Sanskrit. Indian languages like Hindi, Bengali, Sinhala, Oriya and Dravidian languages like Malayalam, Tamil, Telugu, and Kannada borrow many words from Sanskrit. Although recently, \newcite{kanojia2019cognate} perform cognate detection for a few Indian languages, but report results with manual verification of their output. Identification of cognates for improving IR has already been explored for Indian languages \cite{makin2007approximate}. String similarity-based methods are often used as baseline methods for cognate detection and the most commonly used among them is Edit distance based similarity measure. It is used as the baseline in the early cognate detection papers \cite{melamed1999bitext}. Essentially, it computes the number of operations required to transform from source to target cognate.

Research in automatic cognate detection using phonetic aspects involves computation of similarity by decomposing phonetically transcribed words \cite{kondrak2000new}, acoustic models \cite{mielke2012assessing}, phonetic encodings \cite{rama2015comparative}, aligned segments of transcribed phonemes \cite{list2012lexstat}. We study \newcite{rama2016siamese}'s research, which employs a Siamese convolutional neural network to learn the phonetic features jointly with language relatedness for cognate identification, which was achieved through phoneme encodings. Although it performs well on the accuracy, it shows poor results with MRR. \newcite{jager2017using} use SVM for phonetic alignment and perform cognate detection for various language families. Various works on Orthographic cognate detection usually take alignment of substrings within classifiers like SVM \cite{ciobanu2014automatic,ciobanu2015automatic} or HMM \cite{bhargava2009multiple}. We also consider the method of \newcite{ciobanu2014automatic}, which employs dynamic programming based methods for sequence alignment. Among cognate sets common overlap set measures like set intersection, Jaccard \cite{jarvelin2007s}, XDice \cite{brew1996word} or TF-IDF \cite{wu2008interpreting} could be used to measure similarities and validate the members of the set.

\section{Datasets and Methodology}
\label{sec:DnM}

\begin{figure*}[!ht]
\centering
\includegraphics[width=0.8\linewidth]{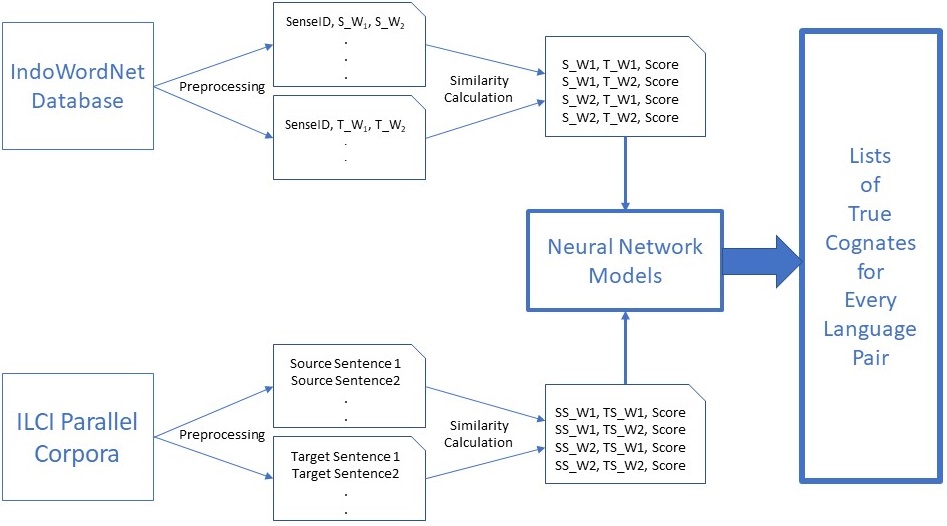}
\caption{Block Diagram for our experimental setup}
\label{fig:bdia}
\end{figure*}

We investigate language pairs for major Indian languages namely Marathi (Mr), Gujarati (Gu), Bengali (Bn), Punjabi (Pa), Sanskrit (Sa), Malayalam (Ml), Tamil (Ta), Telugu (Te), Nepali (Ne) and Urdu (Ur) with Hindi (Hi). We create two datasets as described below for \texttt{<source\textunderscore lang> -<target\textunderscore lang>} where the source language is always Hindi. We describe each step in the subsections below. 

\subsection{Datasets}

\subsubsection*{Dataset 1: WordNet based dataset}

We create this dataset (WNData) by extracting synset data from the IndoWordnet database. We maintain all words, in the concept space, in a comma-separated format. We, then, create word lists by combining all possible permutations of word pairs within each synset. For \textit{e.g.,} If synset ID X on the source side (Hindi) contains words $S_1W_1$ and $S_1W_2$, and parallelly on the target side (other Indian languages), synset ID X contains $T_1W_1$ and $T_1W_2$, we create a word list such as: \\
$S_1W_1,T_1W_1$\\
$S_1W_2,T_1W_1$\\
$S_1W_1,T_1W_2$\\
$S_1W_2,T_1W_2$\\

To avoid redundancy, we remove duplicate word pairs from this list.

\subsubsection*{Dataset 2: Parallel Corpora based dataset}

We use the ILCI parallel corpora for Indian languages \cite{jha2010tdil} and create word pairs list by comparing all words in the source side sentence with all words on the target side sentence. Our hypothesis, here, is that words with high orthographic similarity which occur in the same context window (a sentence) would be cognates with a high probability. Due to the unavailability of ILCI parallel corpora for Sa and Ne, we download these corpora from Wikipedia and align it with the Hindi articles from Hindi Wikipedia. We calculate exact word matches to align articles to each other thus creating comparable corpora and discard unaligned lines from both sides. We, then, create similar word pairs list between Hindi and all the other languages pairs. We removed duplicated word pairs from this list as well and call this data PCData.

\subsection{Script Standardization and Text Normalization}

The languages mentioned above share a major portion of the most spoken languages in India. Although most of them borrow words from Sanskrit, they belong to different language families. Mr, Gu, Bn, Pa, Ne and Ur belong to the Indo-Aryan family of languages; and Ml, Ta, Te belong to the family of Dravidian languages. They also use different scripts to represent themselves textually. For standardization, we convert all the other written scripts to Devanagari. We perform Unicode transliteration using Indic NLP Library\footnote{\url{https://anoopkunchukuttan.github.io/indic_nlp_library/}} to convert scripts for Bn, Gu, Pa, Ta, Te, Ml, and Ur to Devanagari, for both our datasets. Hi, Mr, Sa, and Ne are already based on the Devanagari script, and hence we only perform text normalization for both our datasets, for these languages. The whole process is outlined in Figure \ref{fig:bdia}.

\subsection{Similarity Scores Calculation}
We calculate similarity scores for each word on the source side \textit{i.e.,} Hi by matching it with each word on the target side \textit{i.e.,} Sa, Bn, Gu, Pa, Mr, Ml, Ne, Ta, Te, and Ur.

Since we match the words from the same concept space or the same context window, we eliminate the possibility of this word pair carrying different meanings, and hence \textbf{a high orthographic similarity score gives us a strong indication of these words falling under the category of True Cognates}. For training neural network models, we then divide the positive and negative labels based on a threshold and follow empirical methods in setting this threshold to 0.5 for both datasets\footnote{We ran experiments with 0.25, 0.60, and 0.75 as well, and choose 0.5 based on training performance}. Using 0.5 as threshold, we obtained the best training performance and hence chose to use this as the threshold for similarity calculation. The various similarity measures used are described in the next subsection.

\subsection{Similarity Measures}

\subsubsection*{Normalized Edit Distance Method (NED)}
The Normalized Edit Distance approach computes the edit distance \cite{nerbonne1997measuring} for all word pairs in a synset/concept and then provides the output of probable cognate sets with distance and similarity scores. We assign labels for these sets based on the similarity score obtained from the NED method, where the similarity score is (1 - NED score). It is usually defined as a parameterizable metric calculated with a specific set of allowed edit operations, and each operation is assigned a cost (possibly infinite). The score is normalized such that 0 equates to no similarity and 1 is an exact match. NED is equal to the minimum number of operations required to transform `word a' to `word b'. A more general definition associates non-negative weight functions (insertions, deletions, and substitutions) with the operations.

\subsubsection*{Cosine Similarity (Cos)}
The cosine similarity measure \cite{salton1988term} is another similarity metric that depends on envisioning preferences as points in space. It measures the cosine of the angle between two vectors projected in a multi-dimensional space. In this context, the two vectors are the arrays of character counts of two words. The cosine similarity is particularly used in positive space, where the outcome is neatly bounded in [0,1]. For example, in information retrieval and text mining, each term is notionally assigned a different dimension and a document is characterised by a vector where the value in each dimension corresponds to the number of times the term appears in the document. Cosine similarity then gives a useful measure of how similar two documents are likely to be in terms of their subject matter. This is analogous to the cosine, which is 1 (maximum value) when the segments subtend a zero angle and 0 (uncorrelated) when the segments are perpendicular. In this context, the two vectors are the arrays of character counts of two words. 

\subsubsection*{Jaro-Winkler Similarity (JWS)}

Jaro-Winkler distance \cite{winkler1990string} is a string metric measuring similar to the normalized edit distance deriving itself from Jaro Distance \cite{jaro1989advances}. It uses a prefix scale P which gives more favourable ratings to strings that match from the beginning, for a set prefix length L. We ensure a normalized score in this case as well. Here, the edit distance between two sequences is calculated using a prefix scale P which gives more favourable ratings to strings that match from the beginning, for a set prefix length L. The lower the Jaro–Winkler distance for two strings is, the more similar the strings are. The score is normalized such that 1 equates to no similarity and 0 is an exact match.

\begin{table}[]
\resizebox{0.9\columnwidth}{!}{%
\begin{tabular}{c|c|c|c|c|}
\cline{2-5}
 & \multicolumn{2}{c|}{FFN} & \multicolumn{2}{c|}{RNN} \\ \cline{2-5} 
 & D1 & D2 & D1 & D2 \\ \hline
\multicolumn{1}{|c|}{Hi-Mr} & 69.76 & 85.76 & 74.76 & 89.78 \\ \hline
\multicolumn{1}{|c|}{Hi-Bn} & 65.18 & 81.04 & 69.18 & 86.44 \\ \hline
\multicolumn{1}{|c|}{Hi-Pa} & 73.04 & 78.50 & 76.04 & 83.64 \\ \hline
\multicolumn{1}{|c|}{Hi-Gu} & 61.74 & 79.16 & 69.84 & 89.44 \\ \hline
\multicolumn{1}{|c|}{Hi-Sa} & 61.72 & 85.87 & 68.92 & 91.66 \\ \hline
\multicolumn{1}{|c|}{Hi-Ml} & 56.96 & 74.77 & 66.96 & 79.59 \\ \hline
\multicolumn{1}{|c|}{Hi-Ta} & 55.62 & 61.70 & 65.62 & 68.92 \\ \hline
\multicolumn{1}{|c|}{Hi-Te} & 52.78 & 65.26 & 62.78 & 74.83 \\ \hline
\multicolumn{1}{|c|}{Hi-Ne} & 70.20 & 83.85 & 80.20 & 89.63 \\ \hline
\multicolumn{1}{|c|}{Hi-Ur} & 69.99 & 73.84 & 76.99 & 80.12 \\ \hline
\end{tabular}%
}
\caption{Stratified 5-fold Evaluation using Deep Neural Models on both PCData (D1) and WNData (D2)}
\label{resTable}
\end{table}

\begin{table*}[!ht]
\begin{tabular}{c|c|c|c|c|l|l|l|l|l|l|}
\cline{2-11}
 & \multicolumn{2}{c|}{Corp+WN20} & \multicolumn{2}{c|}{Corp+WN40} & \multicolumn{2}{l|}{Corp+WN60} & \multicolumn{2}{l|}{Corp+WN80} & \multicolumn{2}{l|}{Corp+WN100} \\ \cline{2-11} 
 & FFN & RNN & FFN & RNN & FFN & RNN & FFN & RNN & FFN & RNN \\ \hline
\multicolumn{1}{|c|}{Hi-Mr} & 70.12 & 74.12 & 73.56 & 78.37 & 76.09 & 81.56 & 81.34 & 85.24 & 86.90 & 91.87 \\ \hline
\multicolumn{1}{|c|}{Hi-Bn} & 71.06 & 73.17 & 73.29 & 74.98 & 77.33 & 76.28 & 83.99 & 81.45 & 82.18 & 89.58 \\ \hline
\multicolumn{1}{|c|}{Hi-Pa} & 74.16 & 75.94 & 76.02 & 77.39 & 76.18 & 79.04 & 78.04 & 81.22 & 80.66 & 85.64 \\ \hline
\multicolumn{1}{|c|}{Hi-Gu} & 65.26 & 70.76 & 71.21 & 74.83 & 75.09 & 79.95 & 80.14 & 84.32 & 81.85 & 89.81 \\ \hline
\multicolumn{1}{|c|}{Hi-Sa} & 65.93 & 74.23 & 69.25 & 77.51 & 74.84 & 79.92 & 81.03 & 86.62 & 88.13 & 93.86 \\ \hline
\multicolumn{1}{|c|}{Hi-Ml} & 57.75 & 59.38 & 56.31 & 65.67 & 58.02 & 71.19 & 61.01 & 75.59 & 69.11 & 82.54 \\ \hline
\multicolumn{1}{|c|}{Hi-Ta} & 54.63 & 60.12 & 56.69 & 63.38 & 57.46 & 66.17 & 59.36 & 67.17 & 60.41 & 70.62 \\ \hline
\multicolumn{1}{|c|}{Hi-Te} & 53.21 & 58.18 & 56.19 & 63.90 & 64.15 & 67.70 & 65.19 & 70.65 & 66.10 & 74.92 \\ \hline
\multicolumn{1}{|c|}{Hi-Ne} & 70.78 & 71.23 & 74.30 & 78.11 & 72.19 & 83.20 & 79.70 & 85.01 & 84.69 & 90.95 \\ \hline
\multicolumn{1}{|c|}{Hi-Ur} & 69.94 & 71.25 & 70.01 & 72.35 & 72.03 & 76.59 & 71.07 & 78.27 & 73.99 & 80.99 \\ \hline
\end{tabular}%
\caption{Results after combining chunks of WNData with PCData}
\label{chunkTable}
\end{table*}

\subsection{Models}

\subsubsection{Feed Forward Neural Network (FFN)}
In this network, we deal with a word as a whole. Words of the source and target languages reside in separate embedding space. The source word passes through the source embedding layer. The target word passes through the target embedding layer. The outputs of both embedding lookups are concatenated. The resulting representation is passed to a fully connected layer with ReLU activations, followed by a softmax layer.

\subsubsection{Recurrent Neural Network (RNN)}

\begin{figure}[!ht]
\centering
\includegraphics[width=\linewidth]{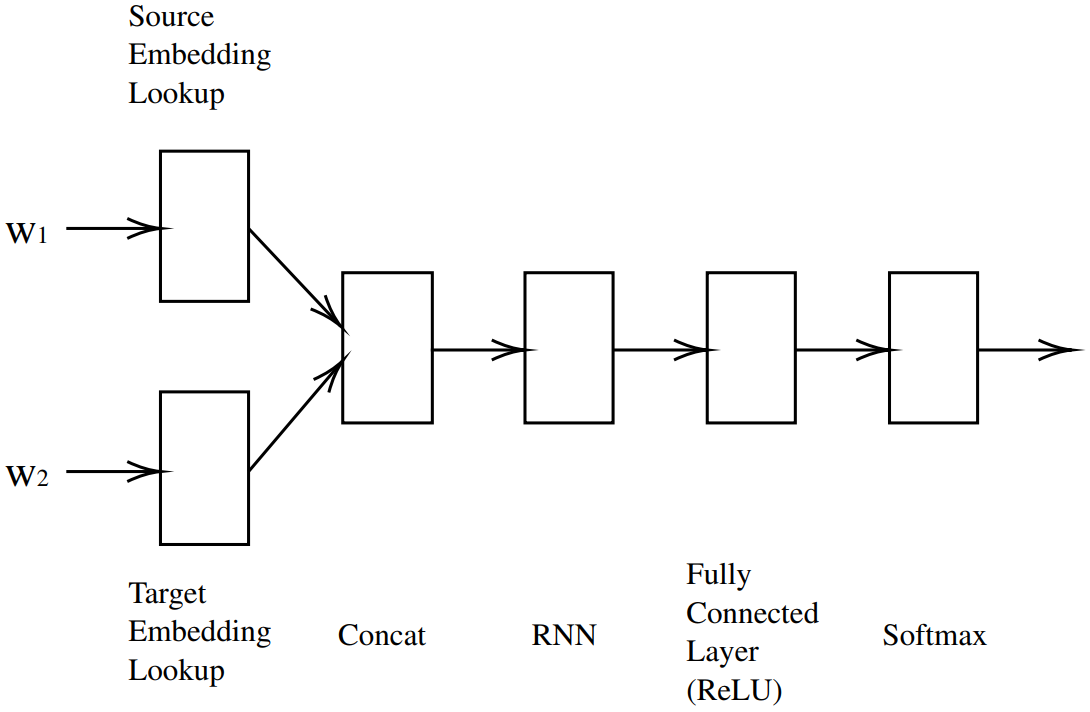}
\caption{Architecture of a Recurrent Neural Network}
\label{fig:RNN}
\end{figure}

In this network (see Figure \ref{fig:RNN}), we treat a word as a sequence of characters. Characters of the source and the target language reside in separate embedding spaces. The characters of the source word are passed through source embedding layer. The characters of the target word are passed through the target embedding layer. The outputs of both embedding lookups are, then, concatenated. The resulting embedded representation is passed through a recurrent layer. The final hidden state of the recurrent layer is then passed through a fully connected layer with ReLU activation. The resulting output is finally passed through a softmax layer.

\section{Results}

\begin{figure*}[!ht]
\centering
\includegraphics[width=\linewidth]{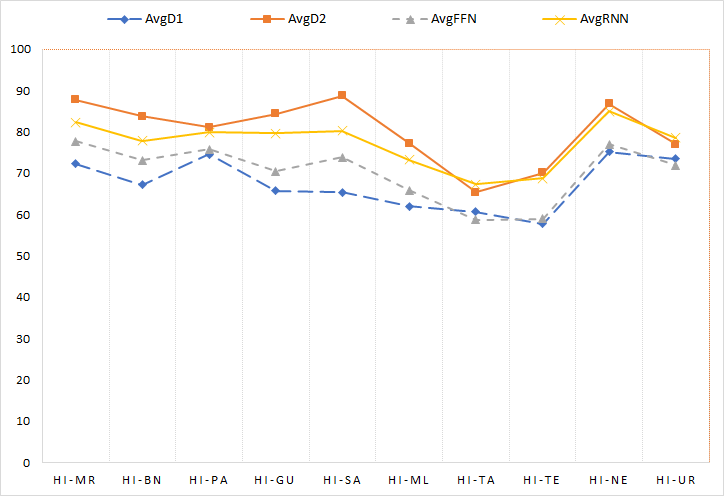}
\caption{Average Results using Neural Network models on both datasets}
\label{fig:f2}
\end{figure*}

We average the similarity scores obtained using the three methodologies (NED, Cos, and JWS) described above, for each word pair, and then use these as training labels for cognate detection models. We obtain results using the networks described above and report them in Table \ref{resTable}. We calculate average scores for both models and both datasets and show the chart in Figure \ref{fig:f2}. We observe that RNN outperforms FFN for both the datasets across all language pairs (see Figure \ref{fig:f2}). We also find that Hi-Sa (see Figure \ref{fig:f2}) has the best cognate detection accuracy among all language pairs (for both RNN and FFN), which is in line with the fact that they are closely related languages when compared to other Indian language pairs. We observe that average scores for WNData are always higher than average scores for PCData for all language pairs (Figure \ref{fig:f2}). Also, in line with our observations above, the overall average of RNN scores for both datasets are even higher than average FFN scores (Figure \ref{fig:f2}).

We perform another set of experiments by combining non-redundant word pairs from both datasets. We add WNData in chunks of 20 per cent to PCData for each language pair and create separate word lists with average similarity scores. We use FFN to train and perform a stratified 5-fold evaluation for each language pair after adding each chunk and show the results in Table \ref{chunkTable}. After evaluating our results for FFN, we perform the same training and evaluation with RNN. \textbf{We observe that adding complete WNData to PCdata improves our performance drastically and given us the best results for almost all cases.} Only in case of Hi-Bn, when using the FFN for training, PCData combined with 80\% WNData performs better than 100\% Data; possibly due to added sparsity of the additional data.  Our hypothesis that adding WNData to PCdata improves the performance holds for all the other cases, including when trained using RNN.

\section{Discussion and Analysis}

\begin{table}[]
\resizebox{0.9\columnwidth}{!}{%
\begin{tabular}{c|c|c|c|}
\cline{2-4}
 & WNPairs & CorpPairs & Matches \\ \hline
\multicolumn{1}{|c|}{Hi-Bn} & 324537 & 505721 & 17402 \\ \hline
\multicolumn{1}{|c|}{Hi-Pa} & 260123 & 465140 & 16325 \\ \hline
\multicolumn{1}{|c|}{Hi-Mr} & 322013 & 555719 & 17698 \\ \hline
\multicolumn{1}{|c|}{Hi-Gu} & 423030 & 542311 & 17005 \\ \hline
\multicolumn{1}{|c|}{Hi-Sa} & 669911 & 248421 & 10109 \\ \hline
\multicolumn{1}{|c|}{Hi-Ml} & 353104 & 315234 & 12392 \\ \hline
\multicolumn{1}{|c|}{Hi-Ta} & 225705 & 248207 & 7112 \\ \hline
\multicolumn{1}{|c|}{Hi-Te} & 369872 & 431869 & 7599 \\ \hline
\multicolumn{1}{|c|}{Hi-Ne} & 191701 & 420176 & 11264 \\ \hline
\multicolumn{1}{|c|}{Hi-Ur} & 99803 & 420176 & 6509 \\ \hline
\end{tabular}%
}
\caption{Total Word Pairs for both datasets and Matches among them}
\label{matchTable}
\end{table}

\begin{table*}[!ht]
\centering
\resizebox{0.95\linewidth}{!}{%

\begin{tabular}{|c|c|c|c|c|c|}
\hline
\textbf{Source Word} & \textbf{Target Word} & \textbf{Meaning} & \textbf{Cos} & \textbf{NED} & \textbf{JWS} \\ \hline
\textit{tadanukool} & \textit{tadanusaar} & accordingly & 0.500 & \textbf{0.571} & 0.482 \\ \hline
\textit{yogadaan karna} & \textit{yogadaan karane} & to contribute & 0.631 & \textbf{0.636} & 0.593 \\ \hline
\textit{duraatma} & \textit{dushtaatama} & evil soul & 0.629 & \textbf{0.700} & 0.648 \\ \hline
\end{tabular}
}
\caption{Manual analysis of the similarity scores}
\label{analysisTable}
\end{table*}

A parallel corpus is a costly resource to obtain in terms of both time and effort. For resource-scarce languages, parallel corpora cannot easily be crawled. We wanted to validate how crucial Wordnets are as a resource and can they act as a substantial dataset in the absence of parallel corpora. In addition to validating the performance of chunks of WNData combined with PCData, we also calculated the exact matches of word pairs from both the datasets and show the results in Table \ref{matchTable}. We observed that Hi-Mr had the most matched pairs amongst all the languages. PCData is extracted from parallel corpora and is not stemmed for root words, whereas WNData is extracted from IndoWordnet and only contains root words. Despite many words with morphological inflections, we were able to obtain exactly matching words, amongst the datasets. WNData constitutes a fair chunk of root words used in PCData as well, and this validates the fact that models trained on WNData can be used to detect cognate word pairs from any standard parallel corpora as well.

It is a well-established fact that Indian languages are spoken just like they are written and unlike their western counterparts are not spoken and spelled differently. Hence, we choose to perform cognate detection using orthographic similarity methods. This very nature of Indian languages allows us to eliminate the need for using aspects of Phonetic similarity to detect true cognates. Most of the Indian languages borrow words from Sanskrit in either of the two forms - \textit{tatsama} or \textit{tadbhava}. When a word is borrowed in \textit{tatsama} form, it retains its spelling, but in case of \textit{tadbhava} form, the spelling undergoes a minor change to complete change. Before averaging the similarity scores, we tried to observe which of the three (NED, JWS, or Cos) scores would perform better for true cognates known to us in \textit{tadbhava} form with minor spelling changes. We analysed individual word pairs from the data and presented a small sample of our analysis in Table \ref{analysisTable}. We observe that NED consistently outperforms Cos and JWS for cognate word pairs and confirmed that NED based similarity is the most suited metric for cognate detection \cite{rama2015comparative}. We also observe that our methodology can handle word pairs without any changes and with minor spelling changes among cognates, the total of which, constitutes a large portion of the cognates among Indian Language pairs.

\section{Conclusion and Future Work}

In this paper, we investigate cognate detection for Indian Language pairs (Hi-Bn, Hi-Gu, Hi-Pa, Hi-Mr, Hi-Sa, Hi-Ml, Hi-Ta, Hi-Te, Hi-Ne, and Hi-Ur). A pair of words is said to be Cognates if they are etymologically related; and True Cognates, if they carry the same meaning as well. We know that parallel concepts, bearing the same sense in linked WordNets, are etymologically related. We, then, use the measures of orthographic similarity to find probable Cognates among parallel concepts. We perform the same task for a parallel corpus and then train neural network models on this data to perform automatic cognate detection. We compute a list of True Cognates and release this data along with the data processed previously. We observe that Recurrent Neural Networks are best suited for this task. We observe that Hindi - Sanskrit language pair, being the closest, has the highest percentage of cognates among them. We observe that RNN, which treats the words as a sequence of characters, outperforms FFN for all the language pairs and both the datasets. We validate that Wordnets can play a crucial role in detecting cognates by combining the datasets for improved performance. We observe a minor, but crucial, increase in the performance of our models when chunks of Wordnet data are added to the data generated from the parallel corpora thus confirming that Wordnets are a crucial resource for Cognate Detection task. We also calculate the matches between word pairs from the Wordnet data and the word pairs from the parallel corpora to show that Wordnet data can form a significant part of parallel corpora and thus can be used in the absence of parallel corpora.

In the near future, we would like to use cross-lingual word embeddings, include more Indian languages, and investigate how semantic similarity could also help in cognate detection. We will also investigate the use of Phonetic Similarity based methods for Cognate detection. We shall also study how our cognate detection techniques can help infer phylogenetic trees for Indian languages. We would also like to combine the similarity score by providing them weights based on an empirical evaluation of their outputs and extend our experiments to all the Indian languages.

\section*{Acknowledgement}

We would like to thank the reviwers for their time and insightful comments which helped us improve the draft. We would also like to thank CFILT lab for its resources which helped us perform our experiments and its members for reading the draft and helping us improve it.

\bibliographystyle{acl2010}
\bibliography{acl2010}
\end{document}